\documentclass[conference]{IEEEtran}
\IEEEoverridecommandlockouts
\usepackage{cite}
\usepackage{amsmath,amssymb,amsfonts}
\usepackage{algorithmic}
\usepackage{graphicx}
\usepackage{textcomp}
\usepackage{xcolor}
\usepackage{fancyvrb}
\def\BibTeX{{\rm B\kern-.05em{\sc i\kern-.025em b}\kern-.08em
    T\kern-.1667em\lower.7ex\hbox{E}\kern-.125emX}}
\begin{document}

\title{Solution to the Non-Monotonicity and Crossing Problems in Quantile Regression\\

}

\author{\IEEEauthorblockN{ Resve A. Saleh}
\IEEEauthorblockA{\textit{Dept. of Electrical and Computer Engineering} \\
\textit{University of British Columbia}\\
Vancouver, Canada \\
res@ece.ubc.ca}
\and
\IEEEauthorblockN{ A.K.Md. Ehsanes Saleh}
\IEEEauthorblockA{\textit{School of Mathematics and Statistics} \\
\textit{Carleton University}\\
Ottawa, Canada \\
esaleh@mathstat.carleton.ca}
}

\maketitle

\begin{abstract}
This paper proposes a new method to address the long-standing problem of lack of monotonicity in estimation of the conditional and structural quantile function, also known as quantile crossing problem.
Quantile regression is a very powerful tool in data science in general and econometrics in particular.
Unfortunately, the crossing problem has been confounding researchers and practitioners alike for over 4 decades. Numerous attempts have been made to find a simple and general solution.
This paper describes a unique and elegant solution to the problem
based on a flexible check function that is
easy to understand and implement in R and Python, while greatly reducing or even
eliminating the crossing
problem entirely. It will be very important in
all areas where quantile regression is routinely used and may
also find application in robust regression, especially in the context of machine learning. From this perspective, we also utilize the flexible check function
to provide insights into the root causes of the crossing problem.
\end{abstract}


\section{Introduction}
Quantile regression, an approach to find different percentiles of
a data set, has been used extensively since the initial concepts
were developed in the late 1970's and early 1980's (Koenker and
Bassett, 1978) \cite{b1}.
However, it exhibits a perplexing feature of producing non-monotonic
behavior that has been dubbed the \textit{crossing problem}.
Those who are new to quantile regression find it somewhat
disconcerting since it violates the basic definition
of percentiles. In fact, it is hard to explain what is the root cause
of the problem, whether it be the kink in the check function, or the lack of a continuous first derivative, or the inability of linear programming
to find a suitable minimum in a sea of hyperplanes defined by the linear constraint functions, or the desired quantile resolution, or perhaps a misspecification of the model. We attempt to answer this question in this paper through the
use of a flexible check function and a series of examples.

In linear models with heteroscedastic errors,
the crossing problem is often exhibited since the 
lines or planes may intersect if there is an unfavorable placement
of observations in a small data set. 
It may also occur in data sets with a large number of
observations, although perhaps not as pronounced depending on the resolution. If the desired resolution is very high, non-monotonic behavior will eventually appear.
However, if the problem can
be reduced or eliminated in most of these cases, 
the results will be more credible and acceptable.

It is important to compartmentalize the 
issue so that an effective solution can be developed and the root causes
of non-monotonicity can be identified. In our view, there are two main components to the crossing problem: 
a systematic component and a random component. 
In the case of quantile regression, we contend that
the systematic error is due to the formulation and method used to obtain the quantile estimates
whereas the random error is due to the data itself or the resolution.
If the systematic error can be removed, then the problem is
reduced to addressing the localized random error.

We introduce an entirely new approach using a flexible check function
to remove the
systematic part of the error. Since the random error tends to be localized,
simple techniques can be used to reduce 
any residual amount of non-monotonicity after the systematic error is removed.
We also provide a new way to evaluate monotonicity of quantile regression estimators.

\section{Continuous Check Function}
For readers already familiar with
quantile regression,  we begin by presenting a new quantile regression function.
We postulate that the root cause of non-monotonicity in quantile regression
originates with the piecewise-linear formulation
of the check function.
If an equivalent continuous check function can be devised, then many of the issues arising
from the original check function can be avoided.
The starting point is our continuous check function given by
\begin{equation}\label{eqn1}
\rho_S(r_i,\tau) =  \frac{1}{20}\text{log}(\text{cosh}(10 r_i))+(\tau-\frac12)r_i
\end{equation} 
where $\tau\in(0,1)$ is the desired quantile and  $r_i$ is the $i$th residual which is $y_i-\textbf{x}_i'\boldsymbol{\beta}(\tau)$ for a linear regression model 
(including the intercept) with error $e_i$ given by 
$$y_i = \textbf{x}_i'\boldsymbol{\beta}(\tau)+e_i\,\,\,\,\,\,i=1,...,n.$$
The quantile regression estimates, $\boldsymbol{\hat\beta}(\tau)$, are obtained by minimizing an associated 
convex loss function which is the conditional quantile function
for each $\tau$ given by
\begin{equation}
Q_S(Y|x,\tau)=\sum_{i=1}^n \rho_S(r_i,\tau)
\end{equation} 
where $n$ is the number of observations or data points in the data set.
At first glance, the equation for $\rho_S(r_i,\tau)$ may not appear to 
be a check function but we demonstrate in a later section that it is, in fact,
a continuous version of the well-known check function. It possesses
continuous 1st and 2nd derivatives (as well as all other derivatives), which offers insights into the
main reasons for the crossing problem. The smoothness and convexity of $Q_S(Y|x,\tau)$ makes it easier to minimize for any given
$\tau$ and the solution is not limited to the set of choices 
available in a linear programming context (i.e. vertices of a polyhedron) but rather
any point in the continuous space. 
Furthermore, any of the nonlinear optimizers in R and Python are
suitable to minimize $Q_S(Y|x,\tau)$.

One caveat is that a function like log(cosh(x)) can overflow if
not implemented correctly. However, it has been used
in machine learning for many years and 
both R and Python have built-in library functions
where this problem is rarely encountered. Also, there will be some
additional runtime cost to compute the function but the results are
greatly improved.
We will spend the next
few sections justifying the use of these two equations as a direct
replacement to the standard quantile regression procedure, and
provide results to support the new approach as well as our conjectures
about the fundamental causes of the crossing problem.

\section{Background}
\subsection{Quantile Regression}
Quantile regression \cite{b10} is a procedure whereby a given set of data
is partitioned into two parts depending on the value of 
a quantile parameter, $\tau \in (0,1)$. It can be viewed as a
collection of related single or multiple linear regressions.
Typically, the collections represent percentiles, deciles
or quartiles based on the selected value of $\tau$ but, in general,
any regular or irregular values of $\tau$ may be specified depending on
the application.
In particular, if $\tau=0.5$, the data is separated into two equal
parts as this is the 50th percentile (or the second quartile). If $\tau=0.75$,
then the data is partitioned into two parts, one which represents $3/4$ of 
the data and the other which represents the remaining $1/4$ of the data.

\begin{figure}[!ht]
    \centering
    \includegraphics[scale = 0.6]{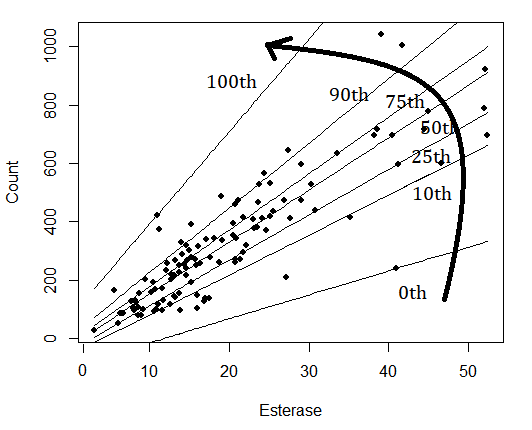}
    \caption{Quantile Regression.}\label{fig:esterase}
\end{figure}

As a concrete example, consider in Fig. \ref{fig:esterase}
which shows a scatter plot of the data and different regression lines
representing different percentiles.
The estimates from a quantile regression produce a separating line that
divides the data based on the specified $\tau$. The 0th percentile
is a line that is located just below all the data points. Similarly, the 100th percentile
is a line that is located just above all the points. The 50th percentile
falls in between and conveniently divides the data in half. Other lines
shown are for the 10th, 25th, 75th and 90th percentile. Note that
as $\tau$ increases from 0 to 1, the slope of the line gets
steeper and steeper as it ``captures" more and more of the points below it.
Essentially, the line sweeps through the points in the radial direction
indicated by the arrow.

One natural approach to monitor the quality of a quantile regression
method is to simply count the number of data points below a given regression
line (or hyperplane in multiple dimensions). The median line (50th percentile)
should result in half the points below it. The 75th percentile
should results in 75\% of the data below that line, and so on.
One rule that should be enforced is that a higher percentile will have
more points below it than a lower percentile. That is, we should not
have more points below the 75th percentile than we do at the 80th percentile.
This would violate the basic definition of percentiles, which is not desirable.
Unfortunately, this type of non-monotonic behavior can occur
with quantile regression, called the crossing problem,
as we will discuss shortly.
\subsection{The Check Function}
\begin{figure}[!ht]
    \centering
    \includegraphics[scale = 0.5]{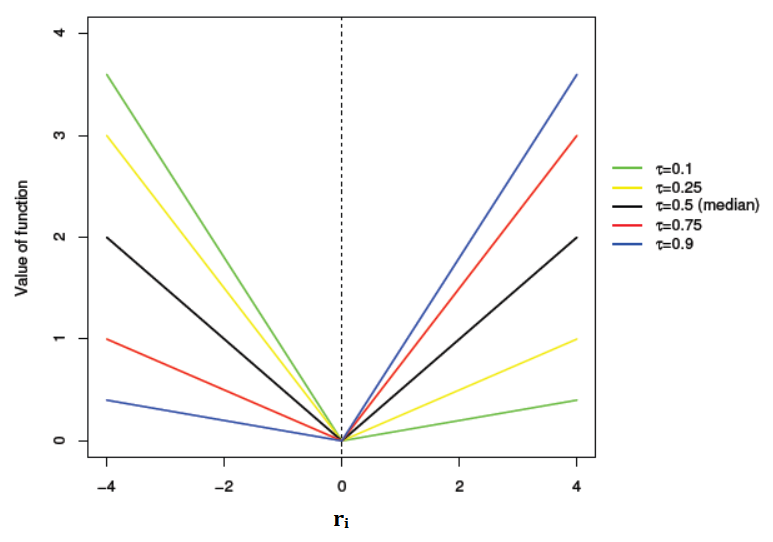}
    \caption{Check function. }\label{fig:check1}
\end{figure}

The question of how to produce the desired regression lines (or hyperplanes)
was first addressed by Koenker and Bassett in their seminal paper \cite{b1} using a formulation
commonly referred to as the check function which can be written concisely as follows:
\begin{equation}\label{eq:check1}
\rho_\tau(r_i,\tau) =
  \begin{cases}
    -(1-\tau) r_i       & \quad \text{if } r_i < 0\\
    \tau r_i  & \quad \text{if } r_i \ge 0
  \end{cases}
\end{equation}
where the scale $ \tau \in (0,1)$ is the quantile of interest 
and $r_i$ is the $i$th residual.
Different regression quantiles represented by 
lines or hyperplanes are obtained by selecting $\tau$ and 
minimizing the conditional quantile function
\begin{equation}
Q_C(Y|x,\tau)=\sum_{i=1}^n \rho_\tau(r_i,\tau).
\end{equation} 
This approach will be referred to as RQ in this paper. A plot of $\rho_\tau(r_i,\tau)$ for different values of $\tau$ is provided in Fig. \ref{fig:check1}.
Suppose we are interested in the median, which is also the 50th percentile and the
2nd quartile. 
We would set $\tau=0.5$ and use the following check function,
\begin{equation}\label{eq:check11}
\rho_\tau(r_i,0.5) =
  \begin{cases}
    -(0.5) r_i       & \quad \text{if } r_i < 0\\
    0.5 r_i  & \quad \text{if } r_i \ge 0.
  \end{cases}
\end{equation}
This is equivalent to the least absolute deviation (LAD) function or the L1 loss
function.
We see from the figure (black line, $\tau=0.5$) that it is symmetric about
$r_i=0$ but
has a kink at 0. This is the same characteristic that causes problems
with the L1 function. Because of this kink, the derivative of this function
is discontinuous at 0. There are mathematical
and numerical issues around using it.
However, for quantile regression, the problem is somewhat exacerbated
as there are many percentiles and quantiles
to consider, other than just the 50th percentile, and they all have
discontinuous 1st derivatives, and undefined 2nd derivatives.

For further illustration, 
suppose we are interested in the 3rd quartile or 75th percentile. Then
we should set $\tau=0.75$ as follows:
\begin{equation}\label{eq:check2}
\rho_\tau(r_i,0.75) =
  \begin{cases}
    -(0.25) r_i       & \quad \text{if } r_i < 0\\
    0.75 r_i  & \quad \text{if } r_i \ge 0.
  \end{cases}
\end{equation}
This produces the asymmetric plot that looks like a \textit{check mark}
as shown in red for $\tau=0.75$ in Fig. \ref{fig:check1}.
This is where the function gets its name; the red lines resemble a
check mark. For other percentiles or quantiles,
it simply requires setting $\tau$ to a different value, as shown in
the figure. Unfortunately, the functions are not smooth and
their derivatives are not continuous.
Every one of them has a kink and a resulting discontinuity
in the first derivative.  When minimizing a loss function constructed
from this check function, one must resort to linear programming techniques
such as the simplex-type or interior point methods to obtain a solution and unfortuantely this presents
a problem as discussed in the next section.
\subsection{The Crossing Problem}
\begin{figure}[!ht]
    \centering
    \includegraphics[scale = 0.5]{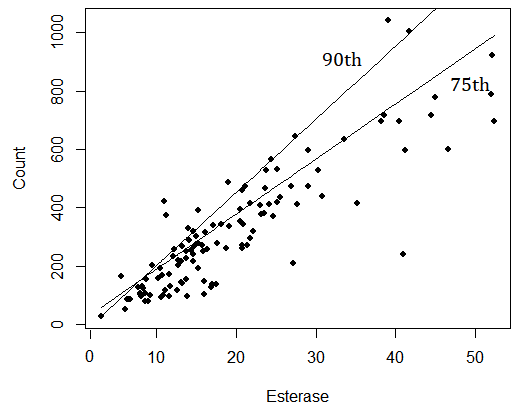}
    \caption{Crossing problem.}\label{fig:esterase2}
\end{figure}

There are countless papers written about the crossing problem and how 
to circumvent it, including \cite{b2}\cite{b3}\cite{b5}\cite{b6}\cite{b7}\cite{b11}\cite{b12}\cite{b8}\cite{b9}\cite{b4}, to name a few. The basic issue is shown in Fig. \ref{fig:esterase2}
and can be contrasted with Fig. \ref{fig:esterase}. We see that the two
lines shown intersect, i.e. cross, and this has implications in the
resulting monotonicity of the percentiles. Not all cases where lines
cross actually present a problem, but the implication is that when
two lines cross in the convex hull of the data, something went wrong.
A higher number of points may be captured by a lower quantile line
as compared to the higher quantile line. Therefore, the
crossing problem is manifested as a non-monotonicity problem
as percentiles increase. But why would this occur?

We suspect that it is due, in part, to the original formulation of the check
function and the techniques used to solve it.
The constraints due to linear programming (a finite set of possible solutions)
and limited options on small data sets as you increase the
percentiles gives rise to a significant part of
the crossing problem and, in particular, the systematic component of the error.
Regions in the neighborhood of $\tau=0$ and $\tau=1.0$ are particularly 
prone to it, although the overall solution may exhibit the non-monotonic
behavior in any region in between. There is also the possibility of infinite
solutions.

Our goal is to find a simple and general solution to avoid crossing.
We emphasize here that the lines and hyperplanes from quantile
regression will always intersect in some way if
they are not parallel. Hence, two lines may cross but the estimated model
can be still be viewed as acceptable if the crossing occurs outside the convex hull \cite{b10}. Lines crossing within the convex hull
often leads to non-monotonicity, but not always,
whereas if non-monotonicity is detected, there is implicitly a crossing problem.
Therefore, we need a way to monitor any non-monotonic behavior
in quantile regression to evaluate the performance of different approaches.

\section{Systematic Regression Quantiles (SRQ)}
\subsection{Continuous L1 Function}\label{AA}
\begin{figure}[!ht]
    \centering
    \includegraphics[scale = 0.5]{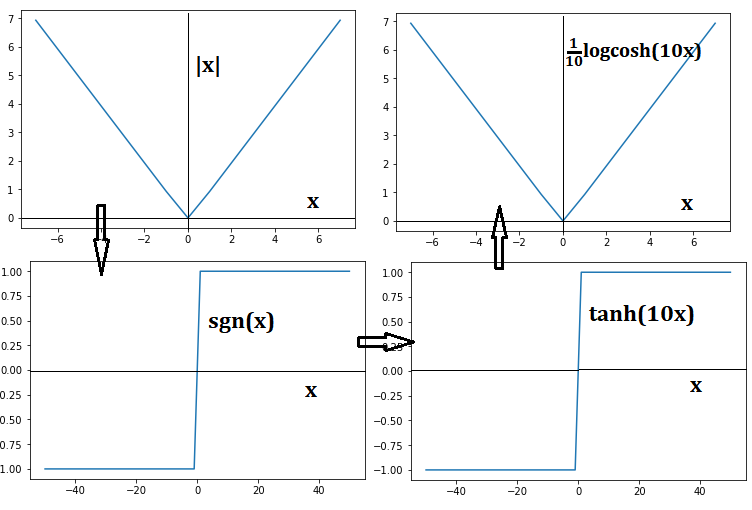}
    \caption{Developing a continuous L1 function.}\label{fig:L1logcosh}
\end{figure}

The starting point for a new approach to quantile regression is
to find a way to construct a continuous function with continuous
derivatives to replace the check function which has discontinuous
derivatives.
Consider first how one might replace the
L1 function with a continuous counterpart. This is shown in Fig. \ref{fig:L1logcosh} using a variable, say $x$. The absolute value function, $|x|$, has a ``V'' shape. When
we take its derivative, we obtain sgn($x$) which is discontinuous at $x=0$.
We can construct a continuous version of it using tanh($10x$). To the naked
eye, they look the same but one is discontinuous and the other is continuous.
Then, by integrating the function, we obtain $\frac{1}{10}$log(cosh($10x$)) which
does not have a kink although it may appear to have one in the figure.
For robust regression, this function can be used as an alternative to LAD.
\subsection{Continuous Check Function}\label{BB}
\begin{figure}[!ht]
    \centering
    \includegraphics[scale = 0.5]{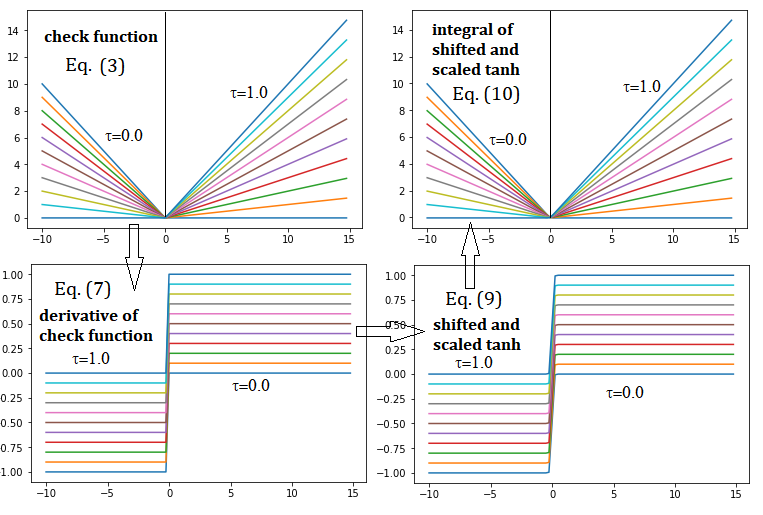}
    \caption{Developing a continuous check function.}\label{fig:check01}
\end{figure}

Now that a procedure is established to create continuous functions
that look like their piecewise-linear counterparts, we merely repeat the
process for the check function. First we take the check function of 
\eqref{eq:check1}
and plot it
as shown in Fig. \ref{fig:check01} in the top left panel
for 11 quantiles (0, 1, and all the deciles in between).
Next we take its derivative to obtain
\begin{equation}\label{eq:check12}
\rho_\tau'(x,\tau) =
  \begin{cases}
    -(1-\tau)        & \quad \text{if } x < 0\\
    \tau   & \quad \text{if } x \ge 0.
  \end{cases}
\end{equation}
This is shown in the bottom left panel. These are all discontinuous
step functions.
We need to create a continuous version of the 
derivative of the check function by first scaling the term tanh($10x$) by
$1/2$ to match its range, that is,
\begin{equation}\label{eqn6}
f(x) =  \frac{1}{2}\text{tanh}(10x).
\end{equation} 
Then we can translate it up or down using $\tau$ by adding $(\tau-\frac12)$,
\begin{equation}\label{eqn7}
f(x,\tau ) =  \frac{1}{2}\text{tanh}(10x)+(\tau-\frac12).
\end{equation} 
This is plotted in the bottom right panel.
Finally, we integrate this function to obtain the continuous check function
of the form given earlier in \eqref{eqn1}:
\begin{equation}\label{eqn8}
F(x,\tau) =  \frac{1}{20}\text{log}(\text{cosh}(10 x))+(\tau-\frac12)x.
\end{equation} 
This is plotted in Fig. \ref{fig:check01} in the top right panel. The use
of \eqref{eqn8} for quantile regression will be called systematic regression quantiles (SRQ) in this paper.

\subsection{Simple Linear Regression Example}
\begin{figure}[!ht]
    \centering
    \includegraphics[scale = 0.6]{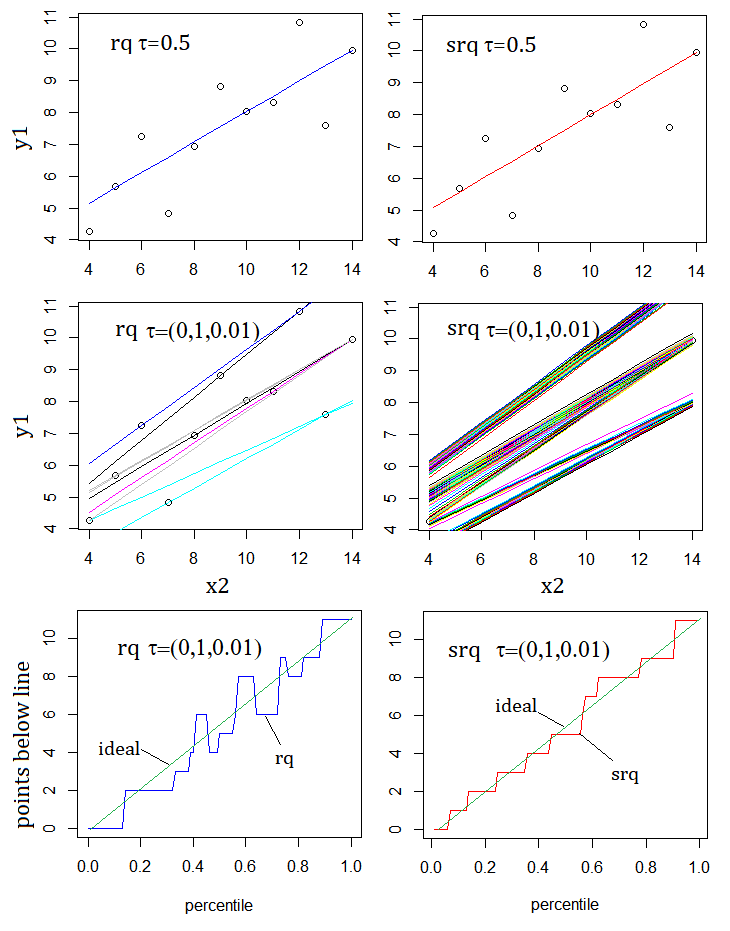}
    \caption{Comparison of RQ and SRQ on anscombe data.}\label{fig:anscombe}
\end{figure}

We will now compare RQ and SRQ on a simple linear example
from the \textit{anscombe} data set in R. Using the $y1$ and $x2$ values,
a scatter plot for the 11 points in the 
data set is shown at the top of Fig. \ref{fig:anscombe}
with the median line ($\tau=0.5$) shown for both RQ and SRQ. 
They are about the same, although on closer inspection, there is a slight difference. The programs
used are \texttt{rq} from the quantreg library in R and \texttt{srq}, our own
R program based on \eqref{eqn8}.

In the middle panels, all 100 regression lines, one-per-percentile, are plotted.
This is a surprising statement since the middle-left panel appears to only
have 10 or 11 lines plotted, whereas the middle-right plot seems to have
many lines plotted (and in fact 100 lines).
It is clear that RQ does not produce 100 distinguishable lines for the 100
percentiles in this case because it is being hindered by the limited number
of options from linear programming. 

On the other hand, SRQ has
the luxury of finding any solution in a continuous space
and therefore produces 100 different lines for the 100 percentiles.
This is the key benefit of SRQ over RQ. In a continuous space, any point
is a potential solution, whereas in the space defined by a set of linear
functions with linear constraints, solutions are permitted only at the vertices
of the constraints, or perhaps an infinite number of solutions are possible which
is also unsatisfactory. This is the major reason for the non-monotonicity
problem encountered in standard quantile regression.

One issue with the linear plots of the middle panels
is that it is difficult to observe
the crossing problem directly because the lines are too close to each other.
We propose a new way to assess the overall performance of a quantile estimator by
simply counting the number of points below a given line. For example, the
0th percentile line should have 0 points below it, while the 100th percentile
line should have all 11 points below it in this example. 
Then, as the percentile increases from 0 to 100, the
number of points should increase monotonically from 0 to 11.
If this does not happen, then the desired monotonicity 
property of percentiles is violated. Such a plot of
``points below line'' against $\tau$ is reminiscent of the Q-Q plot
and can be interpreted in the same way. Hence, the ideal line is a diagonal line
from corner to corner as in a typical Q-Q plot. We will use this type of Q-Q plot
extensively to compare different methods in the remainder of this paper.

The third pair of plots at the bottom of Fig. \ref{fig:anscombe} shows the Q-Q type results
for RQ and SRQ using the quantiles $\tau=0.01,0.02,...,0.99$ which is represented
as $\tau=(0, 1, 0.01)$ in the figure. 
We can observe the significant amount of
non-monotonicity of RQ whereas SRQ closely tracks the ideal solution using an ``uneven staircase''.
The ideal line is the diagonal that connects (0,0) to (1.0,11), the two known solution points, which are interpolated as shown in green. SRQ is monotonic over the entire interval. 
RQ features 3 non-monotonic cycles over the same interval. 

\section{Comparison with RRQ}
A number of techniques for solving the crossing problem have been attempted
in the past with varying degrees of success. Two different approaches
have been taken: (1) Reduce the crossing problem by applying some
constraints or restrictions on the problem, and (2) Apply some smoothing or fitting
method for a number of quantiles over a given interval. See \cite{b5} for
a summary of various methods.

We will first compare SRQ with a well-known approach 
referred to as the Restricted Regression Quantiles (RRQ) method \cite{b2}\cite{b7}.
RRQ as described in \cite{b2} is a three-step method as follows.
Consider the  linear heteroscedastic model given by:
\begin{equation}
y =  x'\beta + (x'\gamma)e
\end{equation}
where $e$ is any error distribution. For RRQ,
first solve the quantile problem for the median case to obtain $\hat\beta$ and $r_i$.
Next, regress $|r_i|$ on $x_i$ to obtain the median regression coefficients
$\hat\gamma$ and the corresponding fitted values which are $s_i=x_i'\hat\gamma$.
Then, finally, we find $c_\tau$ by minimizing $\sum \rho_C(r_i-c s_i)$ over $c$
to produce the $\tau$th hyperplane as $y = x'(\hat\beta+c_\tau \hat\gamma)$,
which is the so-called RRQ estimator.
This approach is implemented in a program called \texttt{rrq} in the Qtools library in R.

\begin{figure}[!ht]
    \centering
    \includegraphics[scale = 0.55]{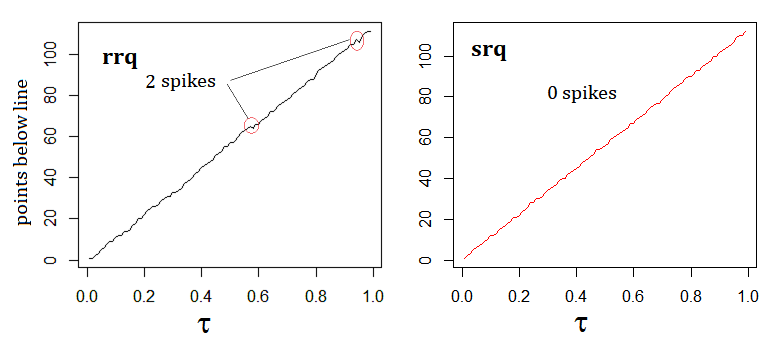}
    \caption{Comparison of RRQ and SRQ on esterase assay data.}\label{fig:ester}
\end{figure}

If we compare RRQ to SRQ on the \textit{esterase assay} data set in R containing 113 
observations and only 1 explanatory variable, we observe in the Q-Q plots of Fig. \ref{fig:ester} 
where RRQ is locally non-monotonic 
in two instances (called spikes in this paper) whereas SRQ is monotonic
for all quantiles shown. Hence, for this data set SRQ is the
preferred option for quantile regression.

\section{Random Error Suppression}
While the SRQ approach 
circumvents the need for linear programming, thus
avoiding the systematic part of the crossing problem, unfortunately 
it does not remove the crossing problem entirely. This is partially
due to the fact that
random errors may arise from the data itself.
However, these random errors tend to be localized by SRQ and can be
removed in a variety of ways. Techniques for 
fitting and local smoothing
are now tractable since most of the errors are single-event errors
(called spikes and pulses) whereas these techniques
are not as effective on the non-monotonic behavior of RQ or RRQ, 
which can be global in nature, especially in small data sets. 

\begin{figure}[!ht]
    \centering
    \includegraphics[scale = 0.6]{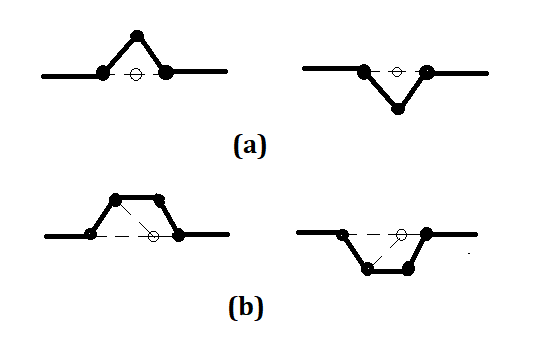}
    \caption{Non-monotonic event suppression: (a) spikes (b) pulses}\label{fig:spikes}
\end{figure}

There are two types of local non-monotonic events that can be handled
easily as shown in Fig. \ref{fig:spikes}. The single-event
positive or negative spikes (solid lines) of Fig. \ref{fig:spikes}(a) can be suppressed (dotted lines)
by examining 2 neighboring quantiles. The single-event positive or negative pulses (solid lines)
of Fig. \ref{fig:spikes}(b) can be converted to spikes, as shown (dotted lines), or removed by considering multiple neighboring
quantiles. The details are not important here 
but the process of
suppressing these events is tractable. If the pulse width exceeds 2 quantiles, more
advanced methods are required.   The 
kernel-based and other fitting approaches \cite{b11} would be effective in those cases.

\begin{figure}[!ht]
    \centering
    \includegraphics[scale = 0.5]{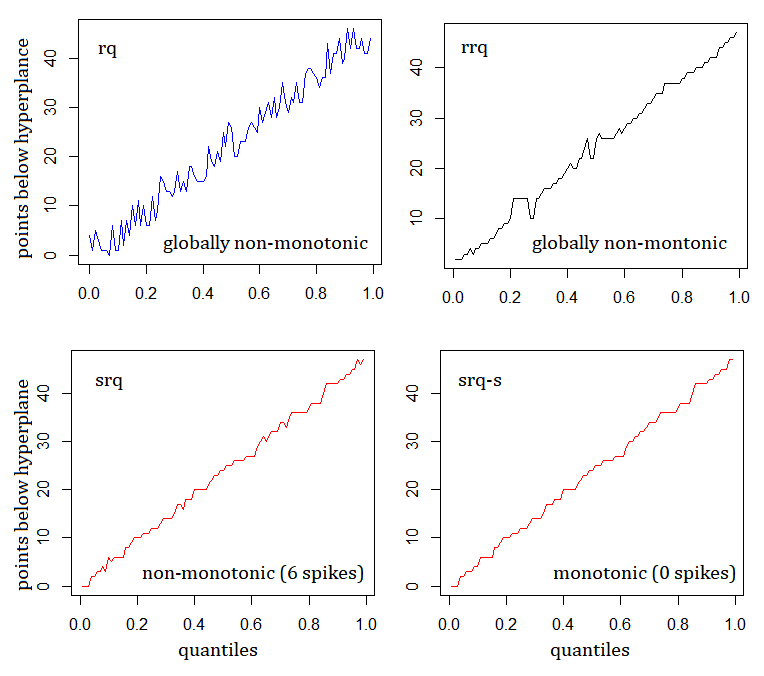}
    \caption{Comparison of RQ, RRQ, SRQ and SRQ-S on swiss data.}\label{fig:swissrq}
\end{figure}

To demonstrate this approach, consider the \textit{swiss} data set with
$n=47$ observations and $p=5$ explanatory variables. We show the Q-Q plots from
quantile regression using RQ, RRQ and SRQ in the top two panels and 
bottom-left panel, respectively, in Fig. \ref{fig:swissrq}. Notice
the severity of the non-monotonic behavior for RQ. It is 
reduced using RRQ, and further reduced using SRQ, but not eliminated.
Both RQ and RRQ exhibit ``global'' non-monotonicity, i.e. something
that cannot be fixed by examining a local region alone.
There are 6 spikes in SRQ that can be removed by simply examining the nearest
quantile neighbors and suppressing them to avoid crossing planes.
This is shown in the bottom-right panel of Fig. \ref{fig:swissrq} which
is referred to as SRQ-S since spike suppression has been applied here.
\begin{figure}[!ht]
    \centering
    \includegraphics[scale = 0.6]{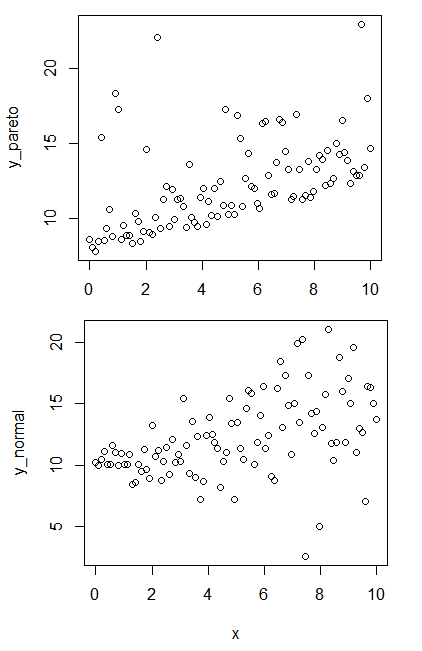}
    \caption{Examples of pareto and normally distributed noise for data with 100 points.}\label{fig:paretonormal}
\end{figure}
\begin{table}[htbp]
\caption{Comparison of rq, rrq and srq on simple linear regression quantiles for different data sizes (pareto).}
\begin{center}\label{tab:pareto}
\begin{tabular}{|c|c|c|c|}
\hline
\textbf{Data}&\multicolumn{3}{|c|}{\textbf{Technique Used}} \\
\cline{2-4} 
\textbf{Points} & \textbf{\textit{rq}}& \textbf{\textit{rrq}}& \textbf{\textit{srq}} \\
\hline
20& 2/1 &  6/1 & 0/0 \\ 
\hline
40& 8/4 &  7/0 & 0/0 \\ 
\hline
60& 10/1 &  1/0 & 0/0 \\ 
\hline
100& 6/0 &  1/0 & 0/0  \\ 
\hline
400& 0/0&  0/0 & 0/0 \\  
\hline
\end{tabular}
\label{tab1}
\end{center}
\end{table}
\begin{table}[htbp]
\caption{Comparison of rq, rrq and srq on simple linear regression quantiles for different data sizes (normal).}
\begin{center}\label{tab:normal}
\begin{tabular}{|c|c|c|c|}
\hline
\textbf{Data}&\multicolumn{3}{|c|}{\textbf{Technique Used}} \\
\cline{2-4} 
\textbf{Points} & \textbf{\textit{rq}}& \textbf{\textit{rrq}}& \textbf{\textit{srq}} \\
\hline
20& 2/3 & 6/1 &  1/0 \\
\hline
40& 3/4 & 10/1 &  1/0 \\
\hline
60& 13/0 & 4/0 & 0/0 \\
\hline
100&  6/0 & 2/0 & 0/0 \\
\hline
400&  0/0 &  0/0 &  0/0  \\
\hline
\end{tabular}
\label{tab2}
\end{center}
\end{table}

In the next series of comparisons, we use will use two cases
that are particularly problematic for standard least squares linear regression.
Synthetic data are created in two ways with 
heteroscedastic normal noise added and 
asymmetric pareto noise added, respectively,
for different numbers of observations (i.e. 20, 40, 60, 100 and 400).
Examples from each set using
100 data points are shown in Fig. \ref{fig:paretonormal}.

The comparison of RQ, RRQ and SRQ involves counting the number of single-event spikes and pulses (of width two or more) over the range of quantiles, 
$\tau = (0,1,0.01)$.
In Tables \ref{tab:pareto} and \ref{tab:normal} comparing \texttt{rq}, \texttt{rrq} and \texttt{srq}, the first number for each is
the number of spikes while the second is the number of pulses. 
We can observe that
\texttt{rq} and \texttt{rrq} have a larger number of such events for small data sets
but it is reduced as the size increases. For \texttt{srq}, the performance
is near-perfect in all cases shown, and can be made perfect using
suppression since they are all single-event spikes, and very few in number.

\section{Smoother Check Functions}
While the results thus far are interesting, it can be
improved further through the use of a smoother check function, as described in
this section.
It is particularly effective in removing non-monotonicity that has to do with
resolution and sensitivity \cite{b5}. In particular, if quantiles
are spaced 0.001 units apart, then a total of 1000 quantiles
are required in the range $(0,1)$. The estimation process may
not have the resolution to produce monotonic behavior using SRQ. 

\begin{figure}[!ht]
    \centering
    \includegraphics[scale = 0.5]{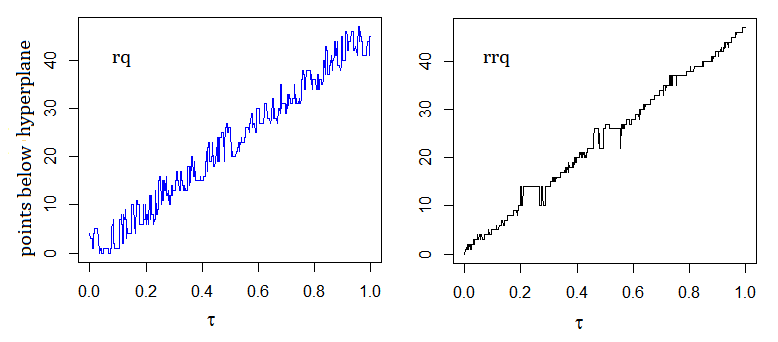}
    \caption{Effect of closely spaced quantiles using  swiss data.}\label{fig:swiss1000rq}
\end{figure}

Consider the
\textit{swiss} data set once again. We show the results of 1000 quantile 
regressions
in Fig. \ref{fig:swiss1000rq}. The effects of the fine resolution
has exacerbated the effects of non-monotonicity compared to Fig. \ref{fig:swissrq}
which was based on 100 quantile regressions using spacings of 0.01.
RQ and RRQ now exhibit much more non-monotonicity.
The root cause of this excessive non-monotonic behavior
is the quantile spacing. The kink is the
source of much of the crossing problem in classical quantile regression.
Although SRQ rounded the kink out 
to make the function continuous, that is still not
sufficient for high resolution quantile regression. 
Therefore, an
adjustment is required to further smoothen the kinks
for the cases where very fine-grain results are desired.

\subsection{Flexible Check Function}
\begin{figure}[!ht]
    \centering
    \includegraphics[scale = 0.75]{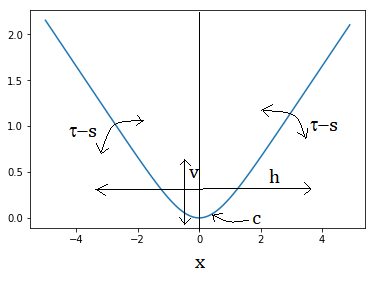}
    \caption{Parameters of flexible check function.}\label{fig:flexrq}
\end{figure}

In this section, we introduce a new approach that
smooths out the loss function.
We described the use of the logcosh function to replicate the original
check function to round out the kink. We use the same type of equation to allow 
more smoothing. Its most general form is the \textbf{key contribution} of our paper as shown in Fig. \ref{fig:flexrq} and given by: 
\begin{equation}\label{eqn20}
F(x,\tau) =  \frac{1}{2c}\text{log}(\text{cosh}(c (x-h)))+(\tau-s)x+v
\end{equation} 
where $c$ is the desired curvature of the function
(i.e. the severity of the kink), $v$ controls vertical shift, $h$ controls the
horizontal shift and $\tau-s$ is
used to produce the desired
slopes on the two sides of the check function itself.

\begin{figure}[!ht]
    \centering
    \includegraphics[scale = 0.53]{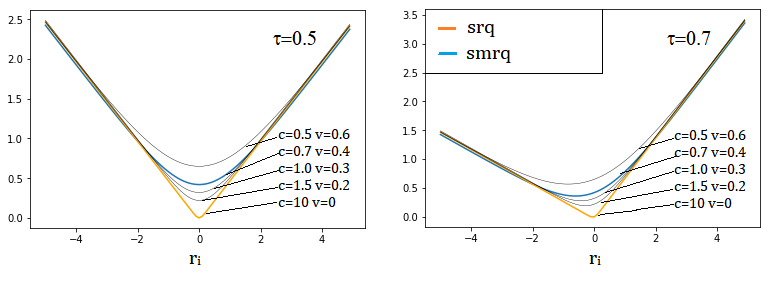}
    \caption{Loss functions for SRQ and SMRQ for $\tau=0.5$ and $\tau=0.7$ cases.}\label{fig:logcoshnew}
\end{figure}

In the case of the continuous check function, we set
$c=10, h=0, s=0.5$ and $v=0$.
However, for a smoother check function, we can use $c=0.7, h=0, s=0.5$ and $v=0.4$
which produces the following 
\begin{equation}\label{eqn21}
\rho_{SM}(r_i,\tau) =  \frac{5}{7}\text{log}(\frac{7}{10}\text{cosh}(r_i))+(\tau-\frac{1}{2})r_i+\frac{2}{5}
\end{equation}
where $r_i$ is the $i$th residual as before.
The method using the above equation
is referred to here as SMRQ for `smoother quantiles'.
Fig. \ref{fig:logcoshnew} shows five different cases of \eqref{eqn20} by varying 
parameters $c$ and $v$ (while holding $s=0.5$ and $h=0$ fixed), including SRQ of \eqref{eqn1} and SMRQ of \eqref{eqn21} for
$\tau=0.5$ and $\tau=0.7$. One can observe the different levels of
smoothness offered by the flexible check function, which can be varied easily
as the need arises.

\begin{figure}[!ht]
    \centering
    \includegraphics[scale = 0.55]{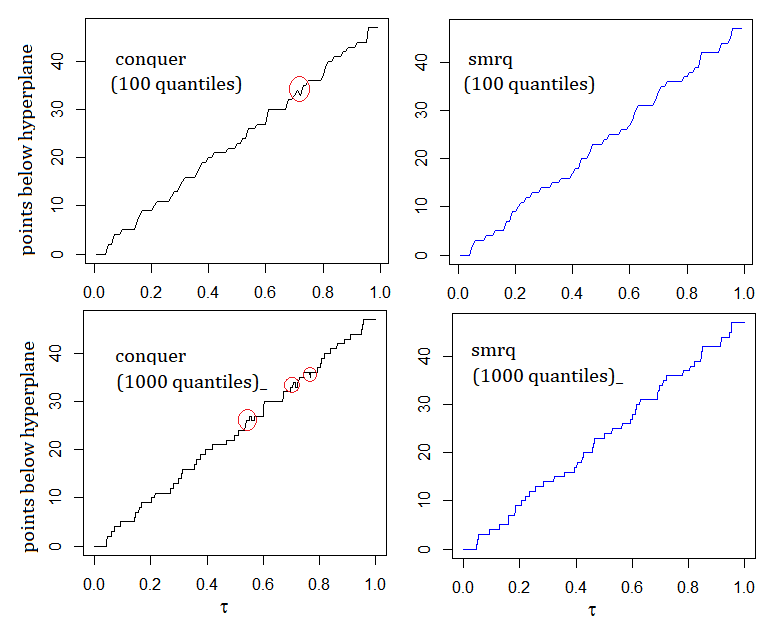}
    \caption{Comparison of conquer and smrq on swiss data. Red indicates a non-monotonic spike event. No pulse events occur.}\label{fig:swiss1000smrq}
\end{figure}

Recently, a technique was published that utilizes convolutional
smoothing applied to the loss function  (He et. al., 2020) and is implemented in the \texttt{conquer} library in R \cite{b4}. 
Convolutional smoothing provides a different approach to reducing the
crossing problem.
A comparison of \texttt{conquer} against \texttt{smrq} is shown in
Fig. \ref{fig:swiss1000smrq}. It is the \textit{swiss} example
with $n=47, p=5$ and quantile spacing of 0.01 and 0.001, respectively. 
We can observe 1 non-monotonic event for the 100 quantile case and
3 non-monotonic events in the 1000 quantile case for \texttt{conquer}, whereas there are none in either case for \texttt{smrq}. We find that 
\texttt{smrq} is generally monotonic in small sample data for
all quartiles, deciles and percentiles.

We now illustrate large-sample data and high-resolution cases using the 
diabetes data with 442 observations
and 10 variables, and the Boston housing data \cite{b5} set with 506 observations
and 15 variables. The number of non-monotonic events (both spikes and pulses) are fewer for \texttt{conquer} compared to \texttt{smrq}. However,
if noise supression is applied to \texttt{smrq}, as described earlier, then \texttt{smrq-s}
is superior to \texttt{conquer}. 
\begin{table}[htbp]
\caption{Comparison of conquer, smrq and smrq-s on diabetes data for different quantile sizes.}
\begin{center}\label{tab:housing}
\begin{tabular}{|c|c|c|c|}
\hline
\textbf{No. of}&\multicolumn{3}{|c|}{\textbf{Technique Used}} \\
\cline{2-4} 
\textbf{Quantiles} & \textbf{\textit{conquer}}& \textbf{\textit{smrq}}& \textbf{\textit{smrq-s}} \\
\hline
100& 0/0 & 0/0 &  0/0 \\
\hline
200& 0/0 & 8/0 &  1/0 \\
\hline
300& 20/1 & 37/5 & 3/0 \\
\hline
400&  35/2 & 90/12 & 12/1 \\
\hline
500&  48/3 &  108/19 &  19/4  \\
\hline
\end{tabular}
\label{tab2}
\end{center}
\end{table}
\begin{table}[htbp]
\caption{Comparison of conquer, smrq and smrq-s on Boston Housing data for different quantile sizes.}
\begin{center}\label{tab:boston}
\begin{tabular}{|c|c|c|c|}
\hline
\textbf{No. of}&\multicolumn{3}{|c|}{\textbf{Technique Used}} \\
\cline{2-4} 
\textbf{Quantiles} & \textbf{\textit{conquer}}& \textbf{\textit{smrq}}& \textbf{\textit{smrq-s}} \\
\hline
100& 0/0 & 0/0 &  0/0 \\
\hline
200& 0/0 & 8/2 &  1/0 \\
\hline
300& 2/0 & 17/3 & 3/1 \\
\hline
400&  6/0 & 38/10 & 4/1 \\
\hline
500&  4/0 &  65/14 &  15/2  \\
\hline
\end{tabular}
\label{tab2}
\end{center}
\end{table}
\section{Conclusions}
In this paper, we have addressed the non-monotonicity problem due to
line and hyperplane crossing in quantile regression. We introduced a unique
flexible check function and used standard nonlinear
optimization to obtain quantile estimates rather than linear programming.
The new method is a simple way of undoing the crossing problem. We compared the baseline method called SRQ with RQ and RRQ to demonstrate
that it is substantially better. Next, we developed a smoothing and suppression
approach, called SMRQ-S, to further reduce the crossing
problem. It was shown to outperform the convolutional smoothing approach used
in \texttt{conquer}.
Finally, we provided cogent explanations of the root causes
of the crossing problem. In summary,
smoothing out the kinks and suppressing spikes reduces the crossing problem significantly in quantile regression.
A portion of our R code is provided in the Appendix using the \textit{swiss}
data set for evaluation purposes.

\section{Appendix: R code for Swiss data}
\begin{Verbatim}[fontsize=\footnotesize]

######################################################
# This contains the Swiss data sets for quantile     # 
# regression comparing rq, rrq, conquer and smrq     #
######################################################

library(quantreg)
library("limma")
library("Qtools")
library("conquer")
swiss <- datasets::swiss
x <- model.matrix(Fertility~., swiss)[,-1]
y <- swiss$Fertility
X <- cbind(x,rep(1,nrow(x)))

minimize.logcosh <- function(par, X, y, tau) {
  diff <- y-(X %*% par) 
  check <- (tau-0.5)*diff+(0.5/0.7)*logcosh(0.7*diff)+0.4
  return(sum(check))
}

smrq <- function(X, y, tau){
  p = ncol(X)
  op.result <- optim(rep(0, p),
            fn = minimize.logcosh,
            method = 'BFGS',
            X = X,
            y = y,
            tau = tau)
    beta <- op.result$par
    return (beta)
}

n <- 99             #choose 99 or 999
p = ncol(X)
vals0 <- rep(0,n)
vals1 <- rep(0,n)
vals2 <- rep(0,n)
vals3 <- rep(0,n)
alphas <- rep(0,n)
for (i in 1:n) {
    alpha = i/(n+1)
    alphas[i] = alpha
    betas <- smrq(X,y,tau=alpha) 
    vals0[i] <- sum(y<(X %*% betas))
    model_qr<-rq(Fertility~.,data = swiss,tau=alpha)
    vals1[i] <- sum(swiss$Fertility < 
              model_qr$fitted.values)
    fitr <- rrq(Fertility ~ ., data = swiss, tau = alpha)
    vals2[i] <- sum(y < predict(fitr))
    alpha1 <- 1.0 - alpha
    fitc <- conquer(x,y,tau = alpha1)
    vals3[i] <- sum(0 <fitc$residual)
}

par(col=1,mfrow = c(2, 2))
plot(alphas, vals1, type = "l", col = 2,main='rq')
plot(alphas, vals2, type = "l", col = 1,main='rrq')
plot(alphas, vals3, type = "l", col = 1,main='conquer')
plot(alphas, vals0, type = "l", col = 4,main='smrq')
\end{Verbatim}
\end{document}